\pdfoutput=1

\documentclass[10pt,twocolumn,letterpaper]{article}

\usepackage{cvpr}
\usepackage{times}
\usepackage{epsfig}
\usepackage{graphicx}
\usepackage{amsmath}
\usepackage{amssymb}
\usepackage{amsfonts} 
\usepackage{pifont}
\usepackage{url}
\usepackage[utf8]{inputenc} 
\usepackage[T1]{fontenc}    
\usepackage{url}            
\usepackage{booktabs}       
\usepackage{amsfonts}       
\usepackage{nicefrac}       
\usepackage{microtype}      
\usepackage{times}
\usepackage{xcolor}
\usepackage{soul}
\usepackage[utf8]{inputenc}
\usepackage[small]{caption}
\usepackage{graphicx} 
\usepackage{wrapfig}
\usepackage{amssymb}
\usepackage{amsmath}
\usepackage{subfigure}
\usepackage{multirow}
\usepackage{float}
\usepackage{authblk}
\usepackage[bookmarks=false]{hyperref}

\def \I {\mathbf{I}}

\def \L {\mathcal{L}}
\def \R {\mathbf{R}}
\def \i {\mathbf{i}}
\def \r {\mathbf{r}}
\def \E {\mathbf{E}}
\def \RR {\mathbb{R}}
\def \FC {\mathbf{FC}}

\makeatletter
\newcommand{\printfnsymbol}[1]{%
  \textsuperscript{\@fnsymbol{#1}}%
}
\makeatother

\cvprfinalcopy 

\def\cvprPaperID{****} 

\ifcvprfinal\fi

\begin{document}

\title{Learning Cross-Modal Embeddings with Adversarial Networks\\ for Cooking Recipes and Food Images}

\author{Wang Hao$^{\dag,}$\thanks{denotes equal contribution} \quad
	Doyen Sahoo$^{\dag,*}$\quad
	Chenghao Liu$^\dag$ \quad
	Ee-peng Lim$^\dag$\quad
    Steven C. H. Hoi$^{\dag,\ddag}$\\
$^\dag$Singapore Management University\quad $^\ddag$Salesforce Research Asia\\
{\{hwang, doyens, chliu, eplim, chhoi\}@smu.edu.sg}
}

\maketitle
\thispagestyle{empty}

\begin{abstract}
Food computing is playing an increasingly important role in human daily life, and has found tremendous applications in guiding human behavior towards smart food consumption and healthy lifestyle. An important task under the food-computing umbrella is retrieval, which is particularly helpful for health related applications, where we are interested in retrieving important information about food (e.g., ingredients, nutrition, etc.). In this paper, we investigate an open research task of cross-modal retrieval between cooking recipes and food images, and propose a novel framework Adversarial Cross-Modal Embedding (ACME) to resolve the cross-modal retrieval task in food domains. Specifically, the goal is to learn a common embedding feature space between the two modalities, in which our approach consists of several novel ideas: (i) learning by using a new triplet loss scheme together with an effective sampling strategy, (ii) imposing modality alignment using an adversarial learning strategy, and (iii) imposing cross-modal translation consistency such that the embedding of one modality is able to recover some important information of corresponding instances in the other modality. ACME achieves the state-of-the-art performance on the benchmark Recipe1M dataset, validating the efficacy of the proposed technique.
   
\end{abstract}


\section{Introduction}
With the rapid development of social networks, Internet of Things (IoT), and smart-phones equipped with cameras, there has been an increasing trend towards sharing food images, recipes, cooking videos and food diaries. For example, the social media platform ``All Recipes" \footnote{https://www.allrecipes.com} allows chefs to share their created recipes and relevant food images. Their followers or fans follow the cooking instructions, upload their pictures for reproducing the same dishes and share their experiences for peer comments. As a result, the community has access to rich, heterogeneous sources of information on food. In recent years, food-computing \cite{min2018survey} has become a popular research topic due to its far-reaching impact on human life, health and well being. Analyzing food data could support a lot of human-centric applications in medicine, biology, gastronomy, and agronomy \cite{min2018survey}. One of the important tasks under the food-computing umbrella is Food Retrieval, i.e., we are interested in retrieving relevant information about a specific food. For example, given a food image, we are interested in knowing its recipe, nutrition content, or calorie information. In this paper, we investigate the problem of cross-modal retrieval between cooking recipes and food images, where our goal is to find an effective latent space to map recipes to their corresponding food images.

\begin{figure}
\begin{center}
\includegraphics[width=0.5\textwidth]{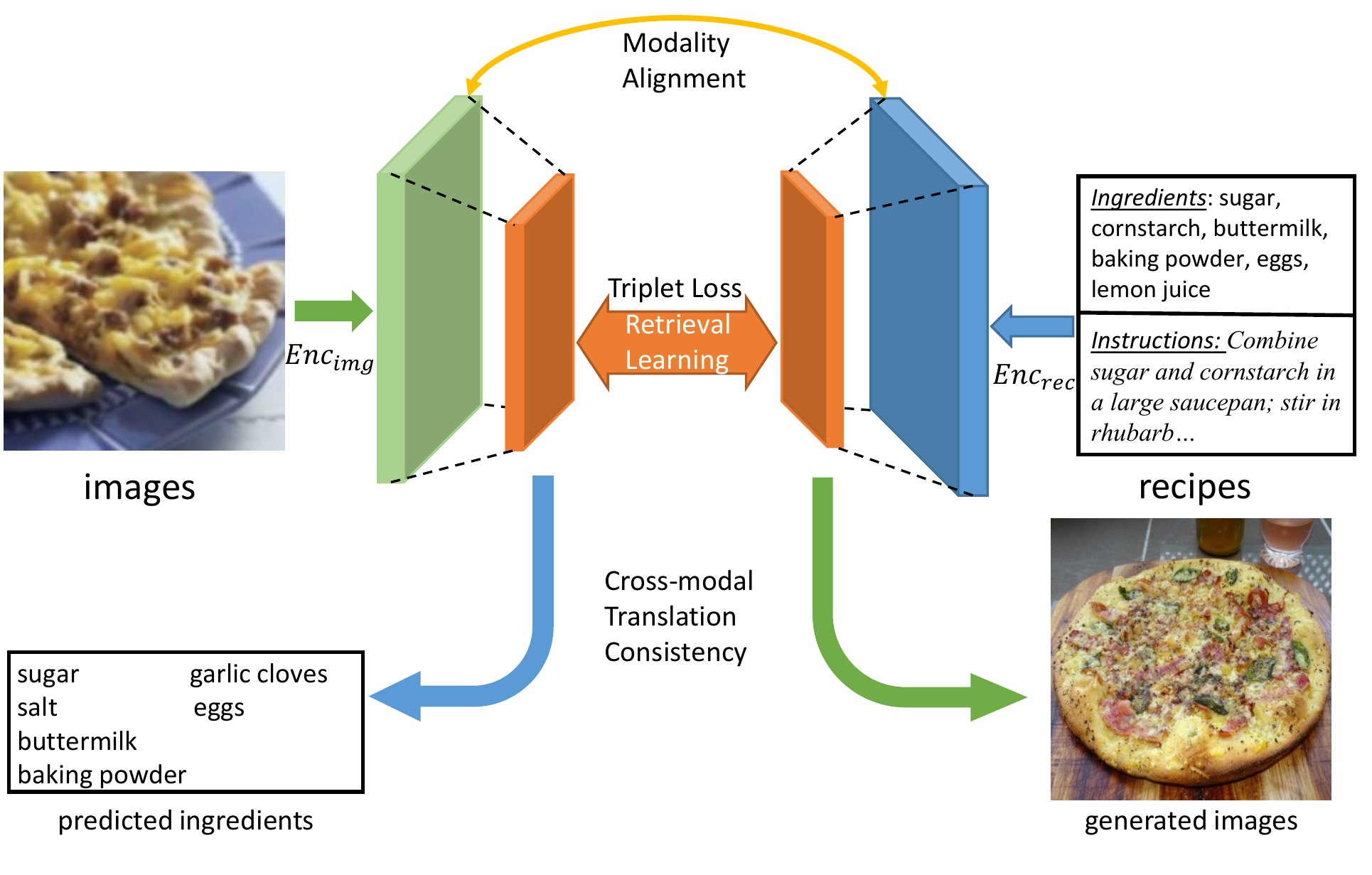}
\end{center}
\vspace{-0.2in}
   \caption{Example of illustrating the idea of Adversarial Cross-Model Embedding (ACME), where the embeddings of cooking recipes and food images are aligned. These embeddings are useful for several health applications from the perspective of understanding characteristics about food, nutrition and calorie intake.}
\label{fig:ACME_concept}
\vspace{-0.1in}
\end{figure}

The idea of cross-modal retrieval in the food domain is to align matching pairs in a common space, so that given a recipe, the appropriate relevant images can be retrieved, or given a food image, the corresponding recipe can be retrieved. Recent efforts addressing the problem for large-scale cross-modal retrieval between cooking recipes and food images \cite{salvador2017learning,carvalho2018cross,chen2018deep} use CNNs \cite{he2016deep} for encoding the food images, and LSTM \cite{hochreiter1997long} to encode recipes, and align the feature vector using common retrieval losses (e.g., pairwise cosine loss and triplet loss). Despite achieving promising performance, there are several critical shortcomings: (i) there can be high variation in images corresponding to one recipe, making it difficult for a triplet loss using a naive sampling strategy to converge quickly; (ii) They do not consider aligning the feature distributions of the two heterogeneous modalities: cooking recipes and food images, which can have very different distributions; and (iii) Cross-Modal mapping using Deep Neural Networks can lead to possible loss of information in the embedding process, and for the existing approaches it is unclear if the embedding of one modality is able to capture relevant semantic information in the other modality.

To address the above limitations, we propose a novel end-to-end framework named \textbf{A}dversarial \textbf{C}ross-\textbf{M}odal \textbf{E}mbedding (ACME). Specifically, we propose an improved triplet loss scheme empowered with hard sample mining, which considerably improves the performance and convergence over a traditional triplet loss. We propose to align the embeddings from the two modalities using an adversarial loss, in an attempt to making the feature distribution of the cooking recipes and food images indistinguishable. We also enforce the cross-modal translation consistency by recovering the relevant information of one modality from the feature embedding of another modality, i.e., we train the model to generate an appropriate image given a recipe embedding, and to predict the ingredients given a food image embedding. 

Figure \ref{fig:ACME_concept} shows an overview of ACME framework, which is end-to-end trainable, and significantly outperforms state-of-the-art baselines for cross-modal retrieval on Recipe1M \cite{salvador2017learning} dataset. We conduct extensive ablation studies to evaluate performance of various components of ACME. Finally, we do qualitative analysis of the proposed method, and visualize the retrieval results. The code is publicly available\footnote{\url{https://github.com/LARC-CMU-SMU/ACME}}.


\section{Related Work}
\subsection{Food Computing}
Food Computing has evolved as a popular research topic in recent years \cite{min2018survey, achananuparp2018eat, achananuparp2018does}. With rapid growth of IoT and the infiltration of social media into our daily lives, people regularly share food and image recipes online. This has given rise to using this rich source of heterogeneous information for several important tasks, and an immense opportunity to analyze food related information. This has far reaching impact to our daily lives, behaviour, health \cite{ofli2017saki,ege2017image} and culture\cite{sajadmanesh2017kissing}. Such analysis could have implications in medicine\cite{farinella2016retrieval}, biology\cite{batt2007food}, gastronomy \cite{mouritsen2017flavour}, agronomy \cite{hernandez2017search} etc. 

There are many ways to analyze food, and many tasks that can be addressed using food computing. One commonly studied task is recognition of food from images. This problem has been studied for years, where large datasets are collected, and prediction models are trained on them. Early approaches used kernel-based models \cite{joutou2009food}, while more recent approaches have exploited deep CNNs \cite{kawano2014food,yanai2015food,liu2016deepfood}. Another task that has received substantial attention is food recommendation. This is a more difficult task than traditional recommendation systems, as a variety of contextual heterogeneous information needs to be integrated to make recommendations. There have been several efforts in literature for food recommendation, including chatbot-based\cite{fadhil2018can}, and dietary recommendation for diabetics\cite{rehman2017diet}. 

Our focus is on another common task: Food Retrieval. We aim to retrieve relevant information about a food item given a query. For example, it can be a difficult task to estimate calorie and nutrition from a food image, but if we could retrieve the recipe and ingredients from an image, the task of nutrition and calorie estimation becomes much simpler. There are several types of retrieval within food computing: image-to-image retrieval \cite{ciocca2017learning}, recipe-to-recipe retrieval \cite{chang2018recipescape}, and cross-modal image-to-recipe and recipe-to-image retrieval\cite{salvador2017learning}. Our work is in the domain of cross-modal retrieval between cooking recipes and food images.

\subsection{Cross-Modal Retrieval}

The goal of cross-modal retrieval is to retrieve relevant instances from a different modality, e.g. retrieving an image using text. The main challenge lies in the media gap ~\cite{peng2018overview}, which means features from different modalities are inconsistent, making it difficult to measure the similarity. 

To solve this issue, many efforts focus on using pairs to learn a similarity or distance metric to correlate cross-modal data \cite{salvador2017learning, socher2014grounded, creswell2016adversarial, wu2010learning}. Apart from metric learning methods, some alignment ideas are also used for cross-modal retrieval: like global alignment~\cite{hotelling1936relations, bach2002kernel, andrew2013deep} and local alignment~\cite{karpathy2014deep, jiang2015deep, niu2017hierarchical}. Canonical Correlation Analysis (CCA) ~\cite{hotelling1936relations} utilizes global alignment to allow the mapping of different modalities which are semantically similar by maximizing the correlation between cross-modal (similar) pairs. In ~\cite{karpathy2014deep}, local alignment is used to embed the images and sentences into a common space. In order to enhance both global and local alignment, \cite{jiang2015deep} learns a multi-modal embedding by optimizing pairwise ranking. Another line of work uses adversarial loss \cite{goodfellow2014generative} which has often been used for alignment in domain adaptation\cite{hoffman2018cycada} and cross-modal retrieval \cite{wang2017adversarial}.

Existing work for large-scale cross-modal retrieval for cooking recipes and food images include JE \cite{salvador2017learning}, which uses a pairwise cosine loss to learn a joint embedding between the two modalities. This method was extended to use a simple triplet loss by AdaMine \cite{carvalho2018cross}. Another approach tried to improve the embedding of the recipes using hierarchical attention \cite{chen2018deep}. Unlike these efforts, our work uses a superior sampling strategy for an improved triplet loss, imposes cross-modal alignment, and enforces cross-modal translation consistency towards achieving more effective and robust cross-modal embedding. 

\section{Adversarial Cross-Modal Embedding}

We now present our proposed Adversarial Cross-Modal Embedding (ACME) between recipes and food images. 

\subsection{Overview}

Given a set of image-recipe pairs $(\i^t, \r^t)$ for $t = 1, \dots, T$, where a food image $\i^t \in \I$ and a recipe $ \r^t \in \R$ (where $\I$ and $\R$ correspond to the image and recipe domains respectively), our goal is to learn embedding functions $\E_V: \I \rightarrow \RR^d$ and $\E_R: \R \rightarrow \RR^d$ which encode the image and the recipe into a $d-$dimensional visual vector and recipe vector, respectively. The embedding functions should be learned so that the embedding of a food image and its corresponding recipe should be close to each other (for $\i^{t1}$ and $\r^{t2}$ where $t1 = t2$), and should be distant from embedding of other images or recipes (for $\i^{t1}$ and $\r^{t2}$ where $t1 \neq t2$). Such an embedding is well suited for retrieval tasks. Note that in our work, the paired instances may have a many-to-one relationship from images to recipes, i.e., we may have many images for a given recipe. Accordingly, we want the embeddings of all the images for a given recipe to be close to the embedding of this recipe, and distant from the embedding of other recipes. 

A simple way to learn this embedding is to use a pairwise cosine loss at the level of feature representation, and use back-propagation to learn the embedding functions \cite{salvador2017learning}. As this loss function may not be ideal for retrieval tasks, triplet-loss was also considered \cite{carvalho2018cross}. However, we hypothesize that both of these approaches suffer from several critical limitations: (i) They give equal importance to all the samples during the optimization of the triplet loss, which may significantly hinder the convergence and generalization, as there could be a high variance among many images corresponding to the same recipe; (ii) Being from heterogeneous modalities, the feature distributions can be very dissimilar, and the existing approaches do not try to align these distributions; and (iii) The embedding functions possibly lose important information during the embedding process, as a result, the embedding of one modality may not effectively capture semantic information in the other modality.

\begin{figure*}
\begin{center}
\includegraphics[scale=0.3]{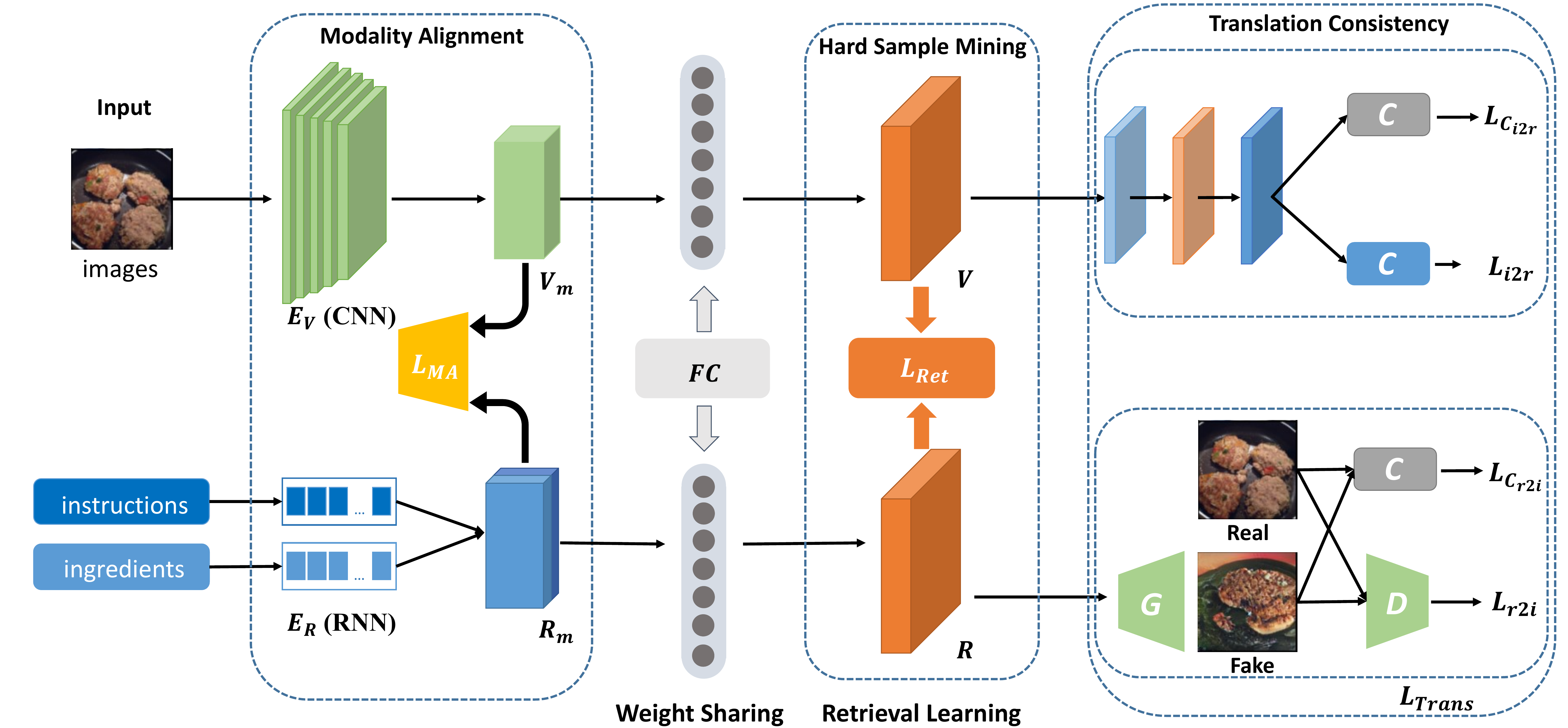}
\end{center}
\vspace{-0.1in}
   \caption{\textbf{ACME: Adversarial Cross-Modal Embedding:} Our proposed framework to achieve effective cross-modal embedding. The food images and recipes are encoded using a CNN and LSTM respectively to give feature embeddings. The feature embeddings are aligned using an adversarial loss $\L_{MA}$ to achieve Modality Alignment. Both image and text embeddings are then passed through a shared FC layer to give the final embedding. This embedding is learned by optimizing $\L_{Ret}$, which uses triplet loss with hard sample mining. This embedding is also optimized to achieve cross-modal translation consistency ($\L_{Trans}$), where the recipe embedding is used to generate a corresponding real food image, and the the food image embedding is used to predict the ingredients in the food item.}
\label{fig:pipeline}
\vspace{-0.1in}
\end{figure*}

To address these issues, we propose a novel end-to-end framework for \textbf{A}dversarial \textbf{C}ross-\textbf{M}odal \textbf{E}mbedding (ACME) between Cooking Recipes and Food Images. To address the first challenge, we propose a new retrieval learning component that leverages a hard sample mining \cite{hermans2017defense} strategy to improve model training and performance of the existing triplet loss. To address the modality-distribution alignment, we use an adversarial loss \cite{goodfellow2014generative} to ensure that features of the embedding functions across different modalities follow the same distribution. Finally, we introduce a novel cross-modal translation consistency component, wherein food images are generated using the embedded recipe features, and the ingredients of a recipe are predicted using the image features. The entire pipeline can be trained in an end-to-end manner with a joint optimization objective. The overall proposed architecture is shown in Figure ~\ref{fig:pipeline}.

More formally, during the feed-forward flow through the pipeline, the food images $\i$ and recipes $\r$ are encoded using a CNN (parameterized by $\E_V$) and an LSTM (parameterized by $\E_R$), respectively. The CNN gives us high-level visual features $V_m \in \RR^d$, and the LSTM gives us high-level recipe features $R_m \in \RR^d$. These high-level features are then passed through a fully-connected layer (parameterized by $\FC$), where both modalities share the same weight \cite{peng2017cm}, with the purpose of correlating the common representation of each modality, and give us the final representation $V$ and $R$ for the visual features and recipe features, respectively. The recipe features $R$ are then used to generate food images, and the visual features $V$ are used to predict the ingredients in the particular instance. This framework is then optimized over three objectives: to achieve a feature representation that is good at retrieval tasks; to obtain a feature representation that aligns the distribution of the two modalities in order to make them modality-invariant; and the features achieve the cross-modal translation consistency. The overall objective of the proposed ACME is given as:
\begin{equation}
\begin{aligned}
\label{eq:main}
\L = \L_{Ret} + \lambda_1 \L_{MA} + \lambda_2 \L_{Trans},
\end{aligned}
\end{equation}
\label{total_formulation}
where $\lambda_1$ and $\lambda_2$ are trade-off parameters. The retrieval learning component $\L_{Ret} (V, R)$ receives the two high-level feature vectors: $V \in \RR^d$ for the image and $R \in \RR^d$ for the recipes, and computes the retrieval loss. The modality alignment component $\L_{MA} (V_m, R_m)$ operates on the penultimate layer features $V_m \in \RR^d $ and $R_m \in \RR^d$, and aims to achieve modality-invariance using an adversarial loss to align the two distributions. The cross-modal translation consistency component $\L_{Trans}$ is further divided into two sub components: recipe2image (generates food image from $R$) and image2recipe (predicts ingredients based on $V$). recipe2image and image2recipe are optimized using an adversarial loss and classification loss respectively. At the end of the training procedure, the feature representations $V$ and $R$ are used for retrieval tasks. 


\subsection{Cross-Modal Retrieval Learning}

After images and recipes are passed through the encoder functions and the weight sharing layers, visual representations $V$ and recipe representations $R$ are obtained. Our goal is to have feature representation $V^{t1}$ and $R^{t2}$ be similar for a given pair ($t1 = t2$) and dissimilar for $t1 \neq t2$. This is done via the usage of the triplet loss \cite{schroff2015facenet}. A triplet comprises one feature embedding as an anchor point in one modality, and a positive and negative feature embedding from another modality. The positive instance corresponds to the one we want it to be similar to the anchor point, and the negative instance should be dissimilar to the anchor point. In our case, we have two types of triplets: one where the visual feature $V$ behaves as the anchor, and another where the recipe feature $R$ behaves as the anchor. The objective is then given as:
\begin{equation}
\begin{aligned}
\L_{Ret} =  & \sum_V\left[d(V_{a}, R_{p})-d(V_{a}, R_{n})+\alpha\right]_+ \; \\
& + \sum_R\left[d(R_{a}, V_{p})-d(R_{a}, V_{n})+\alpha\right]_+,
\end{aligned}
\end{equation}
Our goal is to minimize this objective as:
$$\min_{\E_V, \E_R, \FC} \L_{Ret} $$
Here $d(\bullet)$ is the Euclidean distance, subscripts $a,p$ and $n$ refer to anchor, positive and negative samples, respectively, and $\alpha$ is the margin of error. To improve the convergence of learning, we integrate the hard sample mining strategy into the process of learning with triplet loss. In particular, unlike the traditional approach that simply treats all instances equally important for the triplet construction for a given anchor, the proposed approach gives preference to the most distant positive instance and the closest negative instance during the training procedure.  
\subsection{Modality Alignment}

A major challenge in using encoded features from different modalities is that the distributions of the encoded features can be very different, resulting in poor generalization and slower convergence. A more desirable solution is to align the distribution of the encoded features. We aim to align the distributions of the penultimate layer features $V_m$ and $R_m$. To do so, we use an adversarial loss, where we try to achieve a feature representation such that a discriminator $D_{M}$ cannot distinguish whether the feature representation was obtained from the image or the recipe. We empirically adopt WGAN-GP \cite{gulrajani2017improved} in our experiment. The objective $\L_{MA}$ is given as: 
\begin{equation}
\begin{aligned}
\L_{MA} =  & \mathbb{E}_{\i \sim {p_{image}}} [\log D_M(\E_V(\i)]  + \\
& \mathbb{E}_{\r \sim {p_{recipe}}} [\log(1 - D_M(\E_R(\r)))],
\end{aligned}
\end{equation}
and solved by a min-max optimization as:
$$ \min_{\E_V, \E_R} \max_{D_M} \L_{MA} $$

\subsection{Translation Consistency}

In order to have better generalization, we want the learned embedding of one modality to be able to recover the corresponding information in the other modality, leading to better semantic alignment. This would enforce a cross-modal translation consistency, ensuring that the learned feature representation preserve information across modalities. Specifically, we aim to use recipe features $R$ to generate a food image, and visual features $V$ to predict the ingredients of the recipe. The objective is given as: 
$$\L_{Trans} = \L_{r2i} + \L_{i2r}$$
Below we present the details of each of these losses, which help achieve the cross-modal translation consistency.
\subsubsection{Recipe2Image}

For the recipe2image generation, we have a two-fold goal: the generated images must be realistic, and their food-class identity is preserved. Taking the recipe encoding as input, we use a generator $G_{r2i}$ to generate an image. A discriminator $D_{r2i}$ is then used to distinguish whether the generated image is real or fake. At the same time, a Food-Classifier $C_{r2i}$ is used to predict which food category the generated image belongs to. During training, the discriminator $D_{r2i}$ and classifier $C_{r2i}$ receive both real images and the generated fake images as input. Specifically, we borrow the idea of StackGAN \cite{zhang2017stackgan} to guarantee the quality of generated food images, and the generator $G_{r2i}$ is conditioned on the recipe feature $R = \FC(\E_R(\r))$. The adversarial objective is formulated as:
\begin{equation}
\begin{aligned}
\L_{G_{r2i}} = & \mathbb{E}_{\i \sim {p_{image}}} [\log D_{r2i}(\i)]  + \\
& \mathbb{E}_{\r \sim {p_{recipe}}} [\log(1 - D_{r2i}(G_{r2i}(\FC(\E_R(\r)))))]
\end{aligned}
\end{equation}
While the adversarial loss is able to generate realistic images, it does not guarantee translation consistency. Therefore, we use a food-category classifier which encourages the generator to use the recipe features to generate food items in the appropriate corresponding food-category. The objective is a cross-entropy loss denoted by $\L_{C_{r2i}}$. Combining these two objectives, our optimization is formulated as:
$$\min_{G_{r2i},C_{r2i},\E_R,\FC} \max_{D_{r2i}} \L_{r2i} = \L_{G_{r2i}} + \L_{C_{r2i}}$$

\subsubsection{Image2Recipe}
To achieve the image2recipe translation, we aim to recover the ingredients in the food image. This is done by applying a multi-label classifier on the visual features $V$, which will predict the ingredients in the food image. The multi-label objective is denoted as $\L_{G_{i2r}}$ (as it is in some sense generating the ingredients from a given image).
Like in the case of recipe2image, we also maintain translation consistency by ensuring the features can be classified into the correct food category, using cross-entropy loss $\L_{C_{i2r}}$. The optimization is formulated as:
$$\min_{G_{i2r}, C_{i2r}, \E_I, \FC} \L_{i2r} = \L_{G_{i2r}} + \L_{C_{i2r}}$$
\section{Experiments}


\subsection{Dataset and Evaluation Metrics}
We evaluate the effectiveness of our proposed method on Recipe1M dataset ~\cite{salvador2017learning}, one of the largest collection of public cooking recipe data along with food images. It comprises over 1m cooking recipes and 800k food images. The authors assigned a class label to each recipe based on the recipe titles. This resulted in 1,047 food-classes (which provide relevant information for the cross-modal translation consistency). The authors also identified 4,102 frequently occurring unique ingredients. We adopt the the original data splits ~\cite{salvador2017learning} using 238,999 image-recipe pairs for training, 51,119 pairs for validation and 51,303 pairs for testing. 

We evaluate the model using the same metrics as prior work ~\cite{salvador2017learning,carvalho2018cross}. Specifically, we first sample 10 different subsets of 1,000 pairs (1k setup), and 10 different subsets of 10,000 (10k setup) pairs in the testing set. Then, we consider each item in a modality as a query (for example, an image), and we rank instances in the other modality (example recipes) according to the L2 distance between the embedding of query and that of candidate. Using the L2 distance for retrieval, we evaluate the performance using standard metrics in cross-modal retrieval task. For each test-subset we sampled before (1k and 10k), we compute the median retrieval rank (MedR). We also evaluate the Recall Percentage at top K (R@K), i.e., the percentage of queries for which the matching item is ranked among the top K results.

\begin{table*}[h!]
  \centering
    \caption{\textbf{Main Results.} Evaluation of performance of ACME compared against the baselines. The models are evaluated on the basis of MedR (lower is better), and R@K (higher is better).}
  \begin{tabular}{clrrrrrrrr}
  \toprule
   \textbf{ Size of test-set} && \multicolumn{4}{c}{\textbf{Image to recipe retrieval}} & \multicolumn{4}{c}{\textbf{Recipe to image retrieval} }\\
    \midrule
    \multicolumn{1}{c}{} & \textbf{Methods} & \textbf{medR $\downarrow$} & \textbf{R@1 $\uparrow$} &\textbf{R@5 $\uparrow$}& \textbf{R@10 $\uparrow$} &  \textbf{medR $\downarrow$} & \textbf{R@1 $\uparrow$} & \textbf{R@5 $\uparrow$}& \textbf{R@10 $\uparrow$} \\
    \midrule
	\multirow{6}{*}{{1k}}   
                            &CCA ~\cite{hotelling1936relations}& 15.7 & 14.0 & 32.0 & 43.0 & 24.8 & 9.0 & 24.0 & 35.0 \\
                            &SAN ~\cite{chen2017cross}& 16.1 & 12.5 & 31.1 & 42.3 & - & - & - & -\\
                            &JE ~\cite{salvador2017learning}& 5.2 & 25.6 & 51.0 & 65.0 & 5.1 & 25.0 & 52.0 & 65.0  \\
                            &AM ~\cite{chen2018deep}& 4.6 & 25.6 & 53.7 & 66.9 & 4.6 & 25.7 & 53.9 & 67.1  \\
                            &AdaMine  ~\cite{carvalho2018cross}& 1.0 & 39.8 & 69.0 & 77.4 & 1.0 & 40.2 & 68.1 & 78.7 \\
                            &ACME (ours) & \textbf{1.0} & \textbf{51.8} & \textbf{80.2} & \textbf{87.5} & \textbf{1.0} & \textbf{52.8} & \textbf{80.2} & \textbf{87.6} \\
    \midrule
	\multirow{4}{*}{{10k}} 
	& JE ~\cite{salvador2017learning} & 41.9 & - & - & - & 39.2 & - & - & - \\
	&AM ~\cite{chen2018deep}& 39.8 & 7.2 & 19.2 & 27.6 & 38.1 & 7.0 & 19.4 & 27.8  \\ 
    &AdaMine  ~\cite{carvalho2018cross}& 13.2 & 14.9 & 35.3 & 45.2 & 12.2 & 14.8 & 34.6 & 46.1\\
    & ACME (ours) & \textbf{6.7} & \textbf{22.9} & \textbf{46.8} & \textbf{57.9} & \textbf{6.0} & \textbf{24.4} & \textbf{47.9} & \textbf{59.0}\\
    \bottomrule
  \end{tabular}
  \label{tab:results}
\end{table*}

\subsection{Implementation Details}

Our image encoder $\E_V$ is initialized as a ResNet-50 ~\cite{he2016deep} pretrained on ImageNet ~\cite{deng2009imagenet}, and gives a 1024-dimensional feature vector. A recipe comprises a set of instructions and a set of ingredients. The recipe encoder $\E_R$ is a hierarchical LSTM ~\cite{hochreiter1997long} to encode the instructions (where the word-level embedding is obtained from a pretrained skip-thought algorithm ~\cite{kiros2015skip}), and a bidirectional-embedding to encode the ingredients (where the ingredient embeddings are obtained from word2vec) ~\cite{mikolov2013distributed}. Both embeddings are then concatenated and go through a fully-connected layer to give a 1024-dimensional feature vector. During image generation, from recipe2image, our generator $G_{r2i}$ generated images of size $128\times128\times3$. We set $\lambda_1$ and $\lambda_2$ in Eq. \eqref{eq:main} to be 0.005 and 0.002 respectively. The model was trained using Adam \cite{kingma2014adam} with the batch size of 64 in all our experiments. Initial learning rate is set as 0.0001, and momentum is set as 0.999. 
\subsection{Baselines}

We compare against several state-of-the-art baselines:

\textbf{CCA~\cite{hotelling1936relations}:} Canonical Correlation Analysis is one of the most widely-used  models for learning a common embedding from different feature spaces. CCA learns two linear projections for mapping text and image features to a common space that maximizes feature correlation.

\textbf{SAN~\cite{chen2017cross}:} Stacked Attention Network (SAN) considers ingredients only (and ignores recipe instructions), and learns the feature space between ingredients and image features through a two-layer deep attention mechanism.

\textbf{JE~\cite{salvador2017learning}:} uses pairwise cosine loss to find a joint embedding between different modalities. They attach a classifier to the embedding and predict the food category (1,047 classes).

\textbf{AM~\cite{chen2018deep}:} In addition to triplet loss, AM uses an attention mechanism over the recipe, which is applied at different levels (title, ingredients and instructions). Compared with JE, AM uses additional information from food title.

\textbf{AdaMine~\cite{carvalho2018cross}:} uses a double triplet loss where the triplet loss is applied to both learning a joint embedding and classification of the embedding into appropriate food category. They also integrate the adaptive learning schema which performs adaptive mining for significant triplets.

\subsection{Quantitative Results}

\subsubsection{Main Results}

We show the performance of ACME for cross-modal retrieval against the baselines in Table ~\ref{tab:results}. In retrieval tasks on both 1k, and 10k test datasets, our proposed method ACME outperforms all the baselines across all metrics. On the 1k test dataset, ACME achieves a perfect median rank of 1.0 in both image to recipe and recipe to image retrieval tasks, matching the performance of the current state-of-the-art best method. When we shift to the larger 10k test dataset, where retrieval becomes much more difficult, ACME achieves a Median Rank of less than 6.7 and 6.0 respectively for the two retrieval tasks, which is about 50\% lower than the current state-of-the-art method. The performance of ACME for Recall@K is significantly superior to all the baselines by a substantial margin. On the whole, the performance of ACME is shown to be very promising, beating all the state-of-the-art methods across all the metrics consistently. 

\begin{figure*}
\begin{center}
\includegraphics[width=0.96\textwidth]{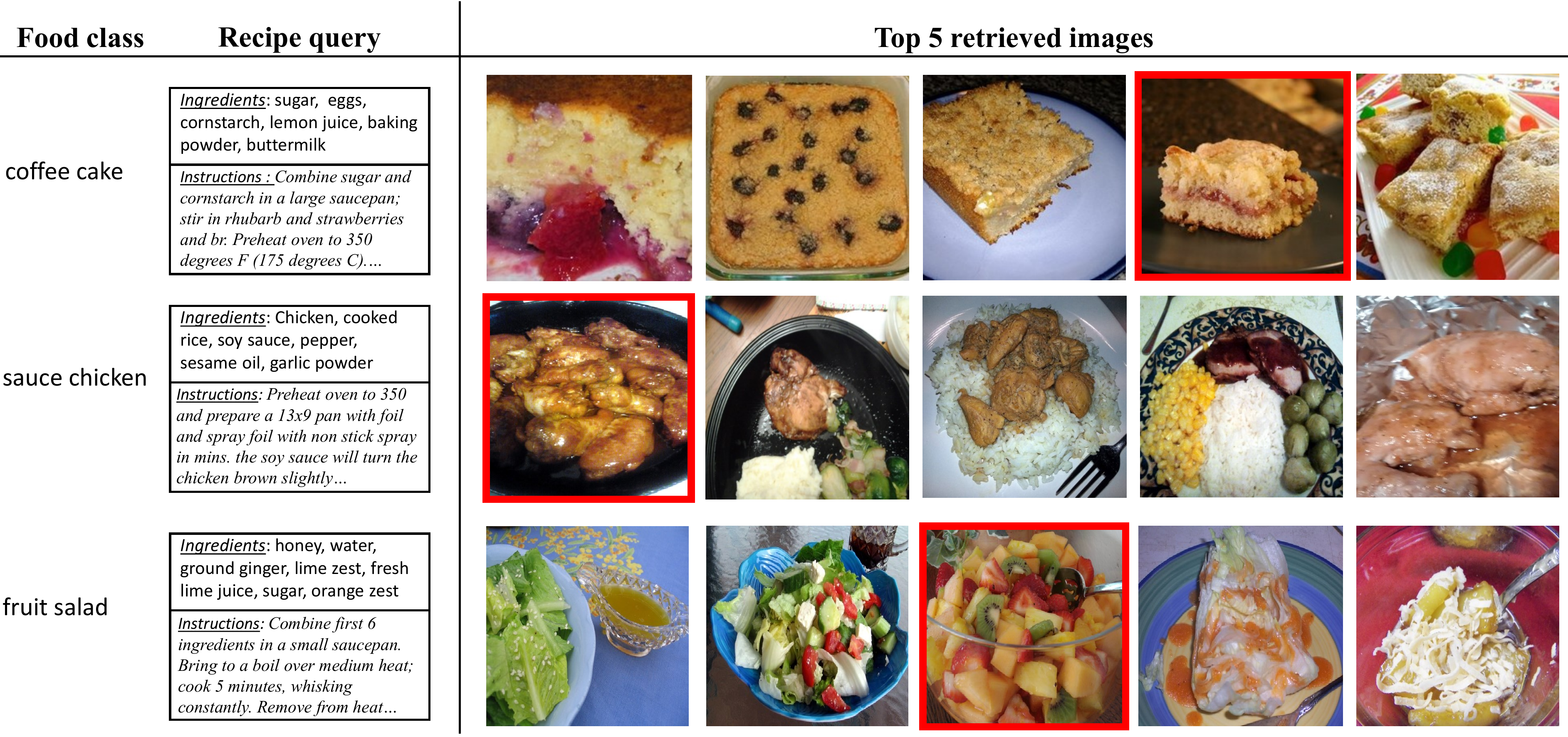}
\end{center}
\vspace{-0.1in}
   \caption{Analysis of Recipe to Image Retrieval on the full 50k test dataset. The first column denotes the food category, and the second column is the query recipe. The top 5 results retrieved by ACME are shown. The best match (i.e., the ground truth) is boxed in red. In most cases, the top retrieved images display similar concepts as the ground truth. }
\label{fig:re2im}
\end{figure*}

\begin{figure*}
\begin{center}
\includegraphics[width=0.96\textwidth]{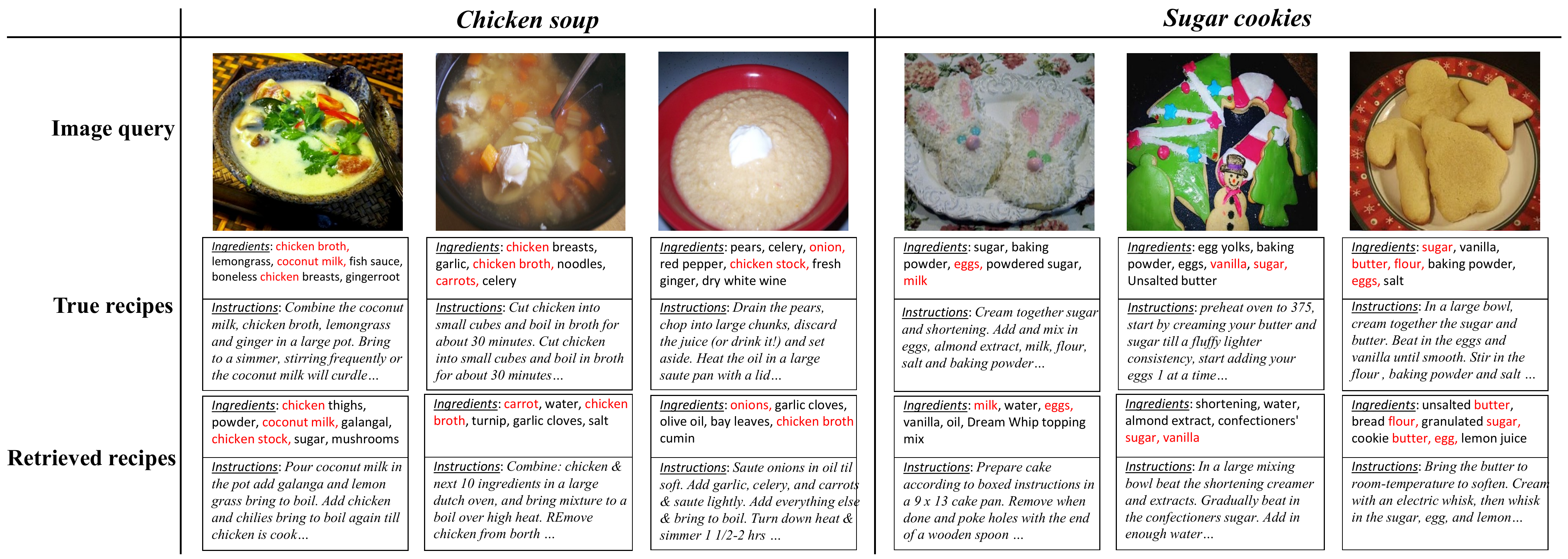}
\end{center}
\vspace{-0.1in}
   \caption{Analysis of Image to Recipe Retrieval on the full 50k test dataset. We consider two food-categories: \emph{chicken soup} and \emph{sugar cookies}. We show 3 different image-queries in each of these categories. Below each image is the true recipe, and below that is the top retrieved recipe by ACME. We can observe that our retrieved recipe has several common ingredients with the true recipe.}
\label{fig:im2re}
\end{figure*}

\subsubsection{Ablation Studies}

We also conduct extensive ablation studies to evaluate the contributions of each of our proposed components in the ACME framework. Specifically, we first evaluate the performance of most basic version of the model with a basic triplet loss \textbf{(TL)}. We incrementally add more components: first, we evaluate the gains from introducing Hard-sample Mining \textbf{(HM)}. Based on this model, we then add the Modality Alignment component \textbf{(MA)}. Based on the resultant model, we then add the cross-modal translation components to measure their usefulness. In particular we want to see the effect of each cross-modal translation \textbf{R2I} and \textbf{I2R}, and their combined effect. We also measure the effect of the full model with and without \textbf{MA}, and finally combine all the components in the ACME framework. We evaluate these components on image-to-recipe retrieval, and the results are shown in Table~\ref{tab:ablation}. In general, we observe that every of the proposed component adds positive value to the embedding model by improving the performance, and all of them work in a collaborative manner to give an overall improved performance.

\begin{table}
  \centering
  \caption{\textbf{Ablation Studies}. Evaluation of benefits of different components of the ACME framework. The models are evaluated on the basis of MedR (lower is better), and R@K (higher is better).}
    \begin{tabular}{lcccc}
    \hline
    \textbf{Component} & \textbf{MedR} & \textbf{R@1} & \textbf{R@5} & \textbf{R@10} \\
    \hline
    \textbf{TL} & 4.1   & 25.9  & 56.4  & 70.1 \\
    \textbf{TL+HM} & 2.0   & 47.5  & 76.2  & 85.1 \\
    \textbf{TL+HM+MA} & 1.9   & 48.0  & 77.3  & 85.5 \\
    \hline
    \textbf{TL+HM+MA+R2I} & 1.4   & 50.0  & 77.8  & 85.7 \\
    \textbf{TL+HM+MA+I2R} & 1.8   & 49.3  & 78.4  & 86.1 \\
    \hline
    \textbf{TL+HM+R2I+I2R} & 1.6   & 49.5  & 77.9  & 85.5 \\
    \textbf{All} & \textbf{1.0} & \textbf{51.8} & \textbf{80.2} & \textbf{87.5} \\
    \hline
    \end{tabular}%
  \label{tab:ablation}%
\end{table}%



\subsection{Qualitative Results}

Here, we visualize some results of ACME. We use the trained ACME model, and perform the two retrieval tasks (Recipe to Image and Image to Recipe) on the full 50k test dataset. Note that this is much harder than the 1k dataset presented in the previous section, and retrieving the correct instance (out of 50k possibilities) is difficult. Our main goal is to show, that despite this difficulty, the top retrieved items appear to be (qualitatively) good matches for the query. We first show the query instance in one modality, and the corresponding ground truth in the other modality. We then retrieve the top results in the second modality, which could show the comparison with the true instance. Even though, often we are not able to obtain the ground truth, ACME retrieves instances that are very similar to this ground truth.

\subsubsection{Recipe to Image Retrieval Results}

We show some examples of recipe to image retrieval in Figure ~\ref{fig:re2im}. We consider 3 food-classes: \emph{coffee cake}, \emph{sauce chicken}, and \emph{fruit salad}, and pick up a recipe to query from each of these classes. ACME is then used to rank the images, and then we look at the top 5 retrieved results. In all cases, the retrieved images are visually very similar to the ground truth image. For example, the images retrieved in case of \emph{coffee cake}, all resemble cakes; images retrieved for \emph{sauce chicken}, all have some form of cooked chicken, couple of them with rice (the ground truth does not have rice, but the ingredients in the query recipe use cooked rice) ; images retrieved for \emph{fruit salad} appear to have fruits in all of them. 
This suggests that the common space learned by our structure has done a good job of capturing semantic information across the two modalities. 

\subsubsection{Image to Recipe Retrieval Results}
We show the image to recipe retrieval results in Figure ~\ref{fig:im2re}. We consider two food categories: \emph{chicken soup}, and \emph{sugar cookies}, and use 3 images in each category as a query image. We retrieve the top recipe for each image query, and compare it with the true recipe corresponding to that image. In the case of \emph{chicken soup}, it is often hard to view the chicken in the food image (as it maybe submerged in the soup), yet the ingredients in the recipe retrieved based on the ACME embedding contains chicken. We also observe several common ingredients between the true recipe and the retrieved recipe for the \emph{sugar cookies} images. Such a performance can have several interesting applications for food-computing. For example, a user can take a food image, retrieve the recipe, and estimate the nutrients and the calories for that food image.

\subsubsection{Cross-Modal Translation Consistency}

An interesting by-product of the proposed ACME framework is the cross-modal translation. These translation components were primarily used to add constraints to obtain a better semantic embedding for our primary retrieval task. We do not claim that these components are robust and generate very good translation (as the food categories are noisy, generating real images is difficult, classification of food into 1,047 categories and multi-label classification of ingredients into over 4,102 categories is very difficult). However, the cross-modal translation can sometimes obtain very interesting results. 

\begin{figure}[h!]
\begin{center}
\includegraphics[width=0.5\textwidth]{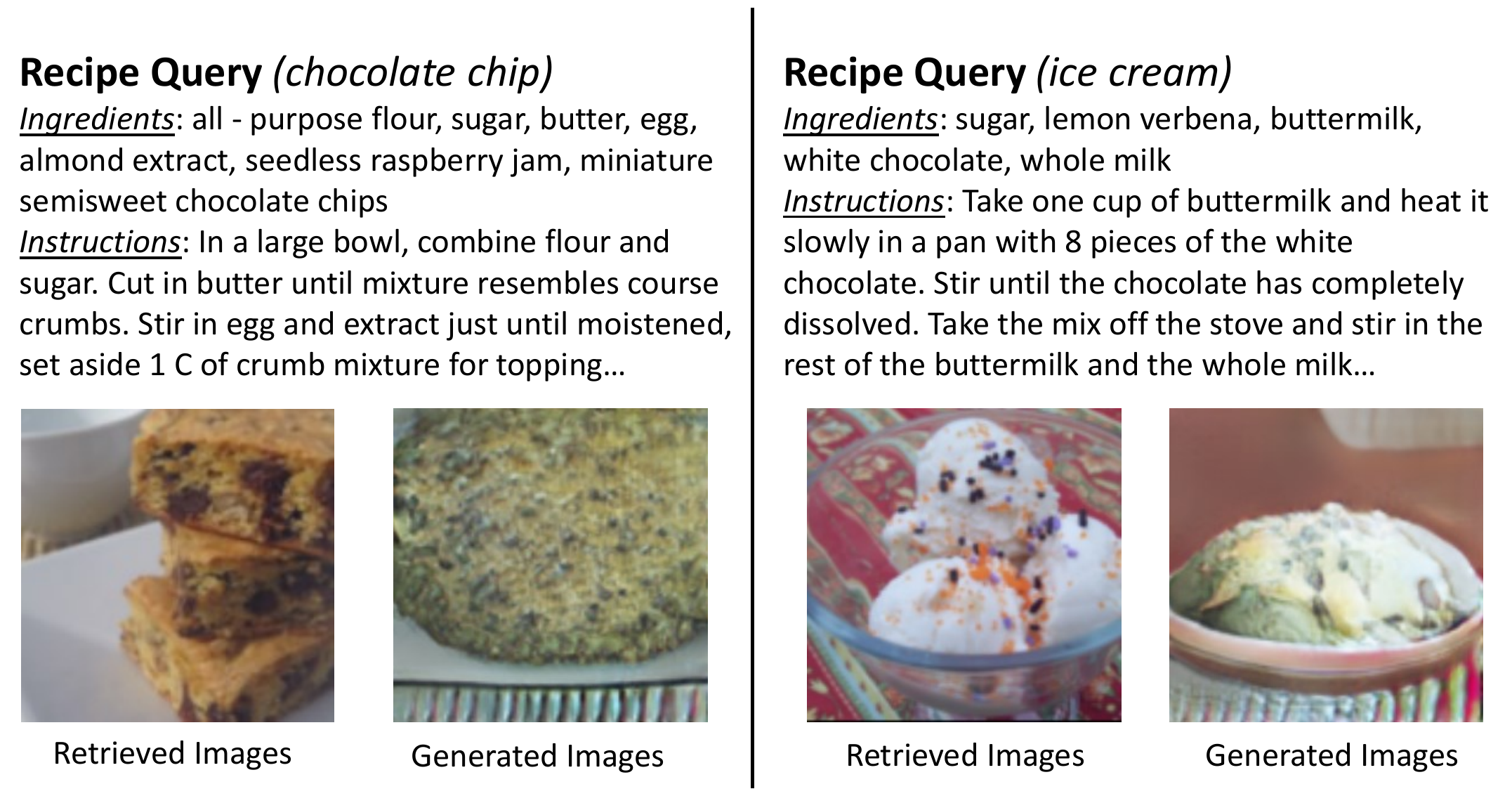}
\end{center}
\vspace{-0.2in}
   \caption{Visualization of the retrieved food images and generated images (generated by r2i translation component) for a query recipe.}
\label{fig:re_gen_im}
\vspace{-0.2in}
\end{figure}

\begin{figure}[h!]
\begin{center}
\includegraphics[width=0.5\textwidth]{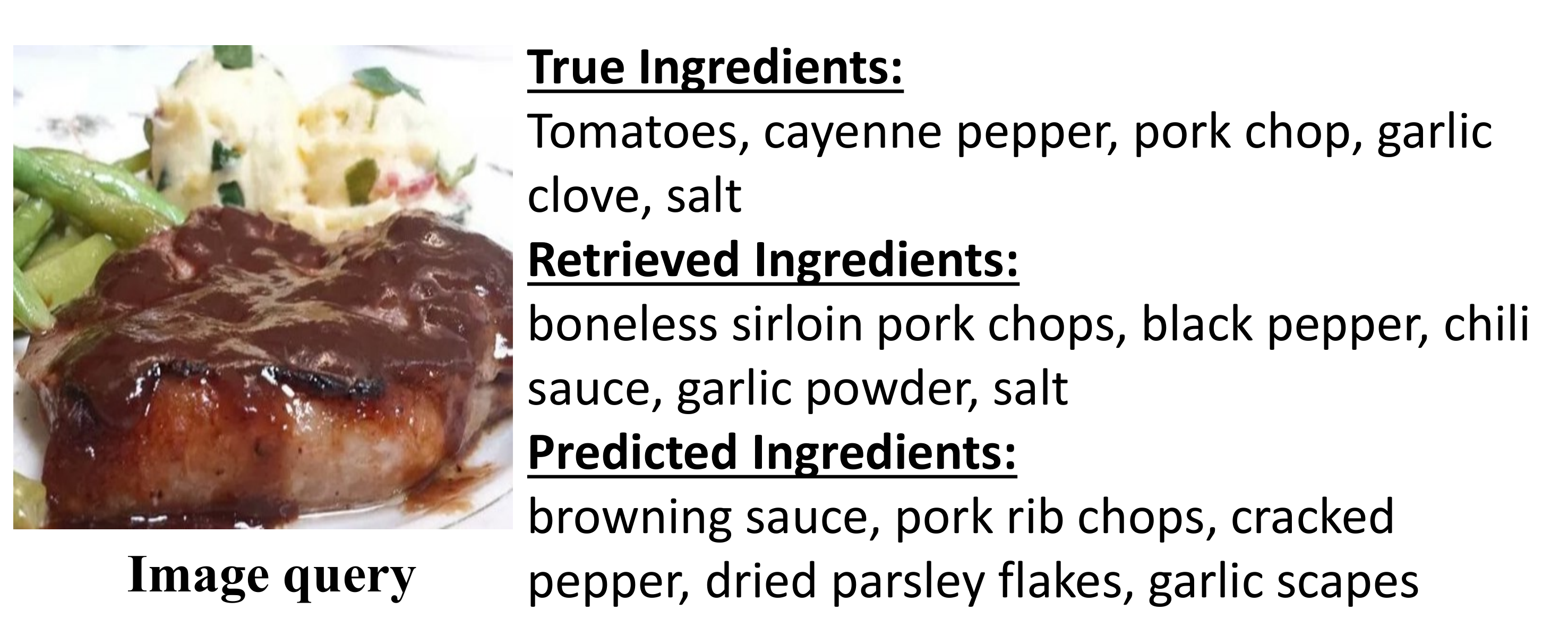}
\end{center}
\vspace{-0.2in}
   \caption{Visualization of the retrieved ingredients and predicted results (predicted by i2r translation component) for query image \emph{pork chops}.}
\label{fig:im_pre_re}
\vspace{-0.2in}
\end{figure}

For example, in Figure ~\ref{fig:re_gen_im} we see a recipe2image translation. Given a recipe query, we retrieve a real food image from the dataset, and compare it with the image generated by the r2i component of ACME. For two instances of \emph{ice cream}, we can see that the generated images show incredible resemblance to the real food images. Similarly, we look at the ingredient prediction from image to see the results of the i2r component in Figure ~\ref{fig:im_pre_re}. The query image is a \emph{pork chop}. We can see several correctly predicted ingredients for the query image. In fact, it is wholly plausible that using the ingredients predicted, it would be possible to cook the dish to obtain a very similar image as the query image. 

\section{Conclusion}
In this paper, we have proposed an end-to-end framework \textbf{ACME} for learning a joint embedding between cooking recipes and food images, where we are the first to use adversarial networks to guide the learning procedure. Specifically, we proposed the usage of hard sample mining, used an adversarial loss to do modality alignment, and introduced a concept of cross-modal translation consistency, where we use the recipe embedding to generate an appropriate food image, and use the food image embedding to recover the ingredients in the food. We conducted extensive experiments, and ablation studies, and achieved state-of-the-art results in the Recipe1M dataset for cross-modal retrieval. 

\section*{Acknowledgments} 
This research is supported by the National Research Foundation, Prime Minister's Office, Singapore under its International Research Centres in Singapore Funding Initiative.

{\small
\bibliographystyle{ieee}
\bibliography{egbib}
}

\end{document}